\pdfoutput=1
\documentclass[11pt]{article}

\usepackage[final]{acl}

\usepackage{times}
\usepackage{latexsym}

\usepackage[T1]{fontenc}
\usepackage[utf8]{inputenc}

\usepackage{microtype}

\usepackage{inconsolata}

\usepackage{graphicx}
\usepackage{cite}
\usepackage{amssymb}
\usepackage{multirow}
\usepackage{xcolor}

\usepackage{algorithm}
\usepackage{algorithmic}
\usepackage{tcolorbox} 
\usepackage{amsmath}
\usepackage{cleveref}       % smart cross-referencing
\usepackage{tcolorbox}

\title{LLaVAC: Fine-tuning LLaVA as a Multimodal Sentiment Classifier}
\author{
 \textbf{Thodsaporn Chay-intr\textsuperscript{1,2}},
 \textbf{Yujun Chen\textsuperscript{3}} \\
 \textbf{Kobkrit Viriyayudhakorn\textsuperscript{2,4}},
 \textbf{Thanaruk Theeramunkong\textsuperscript{1,5}}
\\
 \textsuperscript{1}Intelligent Informatics and Service Innovation Research Center, Thailand \\
 \textsuperscript{2}iApp Technology Co., Ltd., Thailand \\
 \textsuperscript{3}Panasonic Research and Development on Artificial Intelligence (AI), Japan \\
 \textsuperscript{4}Artificial Intelligence Entrepreneur Association of Thailand (AIEAT), Thailand \\
 \textsuperscript{5}Sirindhorn International Institute of Technology, Thammasat University, Thailand \\
\small{
   \texttt{\{t.chayintr, chingyokukun\}@gmail.com, kobkrit@aieat.or.th, thanaruk@siit.tu.ac.th}
}
}

\begin{document}
\maketitle
\begin{abstract}
We present LLaVAC, a method for constructing a classifier for multimodal sentiment analysis. This method leverages fine-tuning of the Large Language and Vision Assistant (LLaVA) to predict sentiment labels across both image and text modalities. Our approach involves designing a structured prompt that incorporates both unimodal and multimodal labels to fine-tune LLaVA, enabling it to perform sentiment classification effectively. Experiments on the MVSA-Single dataset demonstrate that LLaVAC outperforms existing methods in multimodal sentiment analysis across three data processing procedures. The implementation of LLaVAC is publicly available at \url{https://github.com/tchayintr/llavac}.
\end{abstract}

\section{Introduction}
Multimodal Sentiment Analysis (MSA) refers to the process of detecting polarities or attitudes by considering multiple modalities, including images, text, and speech. The polarities (labels) in each modality are typically classified into three categories: positive, negative, and neutral \citep{lopes-etal-2021-automl}. Existing studies in MSA have primarily focused on fusing multiple modalities through complex approaches \citep{cheema-etal-2021-fair,jiang-etal-2020-fusion,li-etal-2022-clmlf}, often leveraging pre-trained models such as BERT \citep{devlin-etal-2019-bert}, RoBERTa \citep{liu-etal-2019-roberta}, and CLIP \citep{radford-etal-2021-learning} to enhance sentiment classification.

Meanwhile, Large Language Models (LLMs) have demonstrated their effectiveness in various language processing tasks, such as text classification \citep{sun-etal-2023-text}. However, their applications have primarily focused on the text domain \citep{naveed-etal-2023-comprehensive}, which may not be well-suited for multimodal tasks, where effective cross-modal understanding and information fusion are essential.

To address this limitation, Multimodal Large Language Models (MLLMs) have been developed, broadening their scope and enhancing their ability to process and integrate multiple modalities \citep{sun-etal-2023-aligning,yang-etal-2023-dawn}. Despite these advancements, their potential within MSA remains underexplored, especially in contexts where MLLMs are employed for classification tasks similar to those tackled by LLMs in the text domain.

Among existing MLLMs, the Large Language and Vision Assistant (LLaVA) \citep{liu-etal-2023-improved,liu-etal-2023-visual-arxiv} stands out as a promising model for multimodal applications due to its ability to process image and text inputs effectively. 
By aligning visual features with language modeling, LLaVA has shown strong zero-shot and fine-tuning capabilities in image-text understanding. This raises a natural question: \textit{Could LLaVA serve as an effective foundation for constructing a multimodal sentiment classifier?}

To address this question, we propose LLaVAC (LLaVA Classifier), a method for fine-tuning LLaVA to serve as a multimodal sentiment classifier. LLaVAC generates sentiment labels for image, text, and their multimodal combination, leveraging LLaVA's strengths in processing multimodal data. This approach aims to enhance MSA capabilities while simplifying the process and reducing the reliance on complex, manual feature engineering. Our contributions are summarized as follows:
\begin{itemize}
    \item We propose LLaVAC, showing that LLaVA, with fine-tuning and prompt design, provides a strong foundation for an MSA classifier.
    \item LLaVAC outperforms eight baseline methods, achieving state-of-the-art performance.
\end{itemize}

\section{Background and Related Work}
\subsection{Multimodal Sentiment Analysis}
MSA is an evolving research field focused on analyzing sentiments in various data types, such as images, videos, audio, and text \citep{cheema-etal-2021-fair}. By combining computer vision, natural language processing, and machine learning, MSA has shown success in areas such as social media, customer service, and product reviews \citep{gandhi-etal-2023-multi}.

Previous work has focused on fusing multiple modalities with pre-trained models to predict a multimodal label that encapsulates sentiment labels for each modality. For example, \citet{cheema-etal-2021-fair} concatenated image and text features initialized from CLIP \citep{radford-etal-2021-learning} and RoBERTa \citep{liu-etal-2019-roberta}, respectively. \citet{wang-etal-2023-exploring} fused both features through Convolutional Neural Networks (CNNs) along with Convolutional Block Attention Module (CBAM) \citep{woo-etal-2018-cbam} where the image and text features were initialized from Residual Networks (ResNet) \citep{he-etal-2015-deep} and BERT \citep{devlin-etal-2019-bert}. 

In addition, recent works such as \citet{sanchez-villegas-etal-2024-improving} introduced auxiliary losses to align and minimize discrepancies between text and image representations, while \citet{chen-etal-2024-holistic} utilized several pre-trained models in combination to enhance sentiment analysis performance.

Although these approaches have achieved state-of-the-art performance, they primarily focus on the multimodal label, often overlooking the contributions of unimodal labels from image and text modalities. Furthermore, existing methods, such as those proposed by \citet{wang-etal-2023-exploring,chen-etal-2024-holistic}, rely on complex fusion strategies, leaving the potential of MLLMs for MSA remains largely unexplored. To the best of our knowledge, the role of MLLMs as a classifier for MSA has not been adequately examined, prompting this study to explore their effectiveness in addressing this task.

\subsection{MLLMs as a Classifier}
Recent advances in LLMs have demonstrated their effectiveness in NLP tasks, inspiring the development of MLLMs capable of integrating multiple modalities, such as images, videos, and audio \citep{naveed-etal-2023-comprehensive}. These models combine the strengths of vision, language, and other modalities, enabling more comprehensive and context-aware understanding across multimodal data. 

However, the use of MLLMs as classifiers for MSA remains relatively unexplored. For instance, \citet{sun-etal-2023-text} utilized RoBERTa and GPT-3 with prompts to build a classifier for unimodal sentiment analysis but did not extend their approach to MSA. To address this limitation, we aim to develop a classifier for MSA by leveraging a MLLM, specifically LLaVA \citep{liu-etal-2023-improved,liu-etal-2023-visual-arxiv}.

\section{Methodology}
\begin{figure}
    \centering
    \scalebox{0.66}{
    \includegraphics{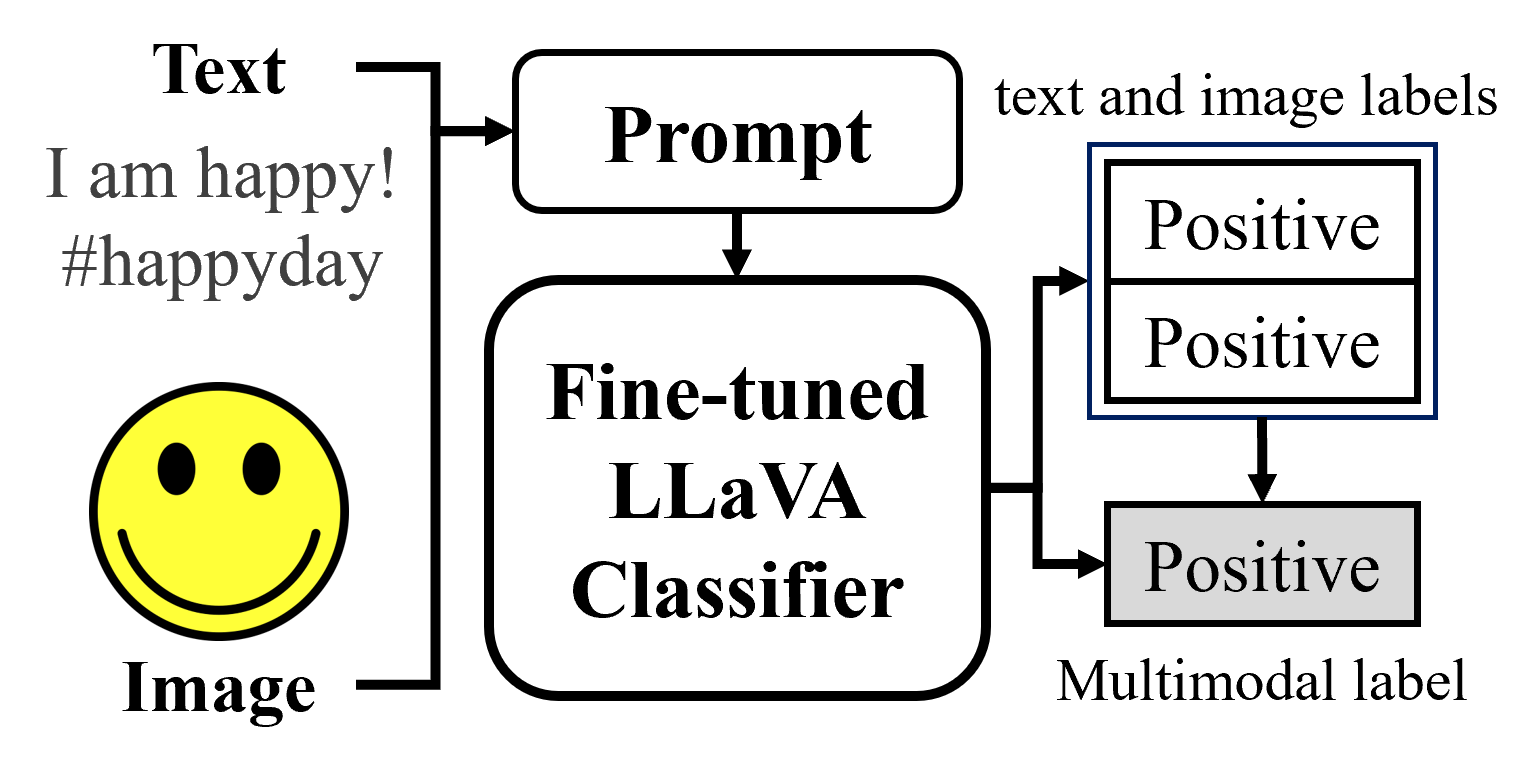}}
    \caption{Our LLaVAC method that utilizes fine-tuned LLaVA classifier to predict image, text, and multimodal labels from a prompt containing image and text data.}
    \label{fig:main-model}
\end{figure}

We propose LLaVAC, a method to build a classifier for MSA, specifically designed for image and text data. This approach classifies sentiment independently for images and text, as well as jointly for their multimodal combination. By fine-tuning LLaVA, our method predicts sentiment labels in a structured, classification-oriented manner, leveraging unimodal labels to complement multimodal label prediction, as illustrated in Figure~\ref{fig:main-model}.

\subsection{Prompt Design}
The core of LLaVAC lies in a structured prompt that explicitly incorporates both image and text modalities during fine-tuning. The prompt is designed to present image, text, and multimodal data alongwith their corresponding sentiment labels in a structured format, restricting classification to sentiment labels (i.e., positive, neutral, and negative).

This setup ensures that the model functions as a classification framework, constraining outputs to predefined sentiment polarities. Additionally, it allows the model to leverage unimodal sentiment labels as context for multimodal label prediction.

\begin{figure}[htbp]
\centering
\begin{tcolorbox}[colframe=black, colback=white!0, boxrule=1pt, arc=0pt, boxsep=4pt, left=4pt, right=4pt, top=2pt, bottom=2pt]
\begin{algorithmic}
\STATE \textbf{Prompt:} 
\STATE Consider the following Image and Text:
\STATE Image: <image>
\STATE Text: RT @babeshawnmendes: "that was really energetic"
\STATE
\STATE Classify the sentiment labels (negative, neutral, positive) for the Image and Text labels separately.
\STATE Finally, jointly analyze and classify the multimodal label for both the Image and Text.
\STATE Provide a short answer with 3 labels for the Image, Text, and Multimodal labels, respectively.
\STATE
\STATE \textbf{Response:}
\STATE positive, positive, positive
\end{algorithmic}
\end{tcolorbox}
\caption{Example prompt with its response, including image, text, and multimodal labels, respectively. <image> denotes the LLaVA-compatible image data, used for fine-tuning the model.} 
\label{fig:prompt-example}
\end{figure}

Figure~\ref{fig:prompt-example} illustrates the prompt and response, directing the model to classify image and text labels independently, followed by a joint analysis to determine the multimodal label. This design explicitly represents sentiment labels for each modality, guiding the model to learn and predict sentiment in a structured and classification-oriented manner.

To ensure compatibility with LLaVA’s architecture, we designed the prompt-response format to follow its inherent processing and fine-tuning requirements.\footnote{\url{https://github.com/haotian-liu/LLaVA/blob/main/docs/Finetune_Custom_Data.md}} This alignment ensures each image-text pair conforms to LLaVA’s prompt format.

\subsection{LLaVA Classifier}
We build LLaVAC by fine-tuning LLaVA using our designed prompts and responses. The fine-tuned model is used during inference in a zero-shot classification setting to process prompts and generate structured responses, as shown in Figure~\ref{fig:prompt-example}. 

In this configuration, the model generates shorter, more focused outputs compared to standard LLMs. These outputs are designed to provide sentiment labels for the image, text, and their multimodal combination, serving as a classifier for MSA.

\section{Experiments}

\subsection{Dataset}
We selected the MVSA-Single dataset\footnote{\url{https://mcrlab.net/research/mvsa-sentiment-analysis-on-multi-view-social-data}} to evaluate our approach. It consists of image-text pairs from Twitter, with sentiment labels assigned independently to the image and text by a single annotator.

We processed the dataset using three procedures. First, we divided it into 10 splits\label{label:10-fold-splits}, each containing training, validation, and test sets, as described by \citet{cheema-etal-2021-fair} for fairness evaluation. Second, we randomly split the dataset into training, validation, and test sets in an 8:1:1 ratio, following \citet{xu-etal-2017-multi,wang-etal-2023-exploring}, and set a random seed of 42 for reproducibility. Third, we used the splits provided by \citet{zhang-etal-2023-provable}.

\subsection{Experimental Settings}
We utilized LLaVA (v1.5-7b)\footnote{\url{https://huggingface.co/liuhaotian/llava-v1.5-7b}} as the base model and fine-tuned it using the LLaVA hyperparameters\footnote{\url{https://github.com/haotian-liu/LLaVA}} with LoRA \citep{hu-etal2021-lora}, as suggested by \citet{liu-etal-2023-improved,liu-etal-2023-visual-arxiv}. The model was fine-tuned for a single epoch to evaluate LLaVA adaptability and capability for MSA with minimal training. This setup aims to simplify the fine-tuning process and reduce the need for complex, manual feature engineering, making it practical for resource-intensive models. Table~\ref{tab:hyper-param} summarizes the essential hyperparameters used for fine-tuning LLaVAC.

\begin{table}[h!]
\centering
\scalebox{0.87}{
\begin{tabular}{lr}
\hline
\textbf{Parameter} & \textbf{Value} \\ \hline
Epoch          & 1             \\ 
Batch size      & 32           \\ 
Initial learning rate          & 2e-5              \\ 
Seed          & 42              \\ 
\hline
Temperature          & 0.01           \\ 
LoRA rank          & 128           \\ 
\hline
\end{tabular}}
\caption{Essential hyperparameters for LLaVAC.}
\label{tab:hyper-param}
\end{table}

Unlike previous works such as \citet{cheema-etal-2021-fair}, which removed hashtags and links to optimize test set performance, our approach retains these elements in the text samples during both fine-tuning and testing. By retaining hashtags and links, we aim to preserve the full context of the text data.

We fine-tuned LLaVA using only the training set and evaluated it on the test set, differing from prior studies \citep{xu-etal-2017-multi,cheema-etal-2021-fair,li-etal-2022-clmlf,wang-etal-2023-exploring}, which included a validation set in their workflow. Additionally, we used Apache Spark\footnote{\url{https://spark.apache.org}} for dataset processing, ensuring efficient, scalable, and adaptable data handling.

\subsection{Evaluation Metrics}

We evaluated our approach using accuracy and weighted F1-scores, following the methodology in \citet{cheema-etal-2021-fair}. Both metrics were computed based solely on the multimodal label to assess overall performance. The evaluation was conducted across the three dataset processing procedures described in Section~\ref{label:10-fold-splits}. For the 10-fold split procedure, we calculated the average scores across all splits to ensure a comprehensive evaluation. 

\subsection{Results}
Table~\ref{tab:main-result} presents the evaluation results, comparing LLaVAC with baseline models based on their reported scores. Our model consistently outperforms all baselines across all evaluation metrics. This demonstrates that fine-tuning LLaVA is effective for MSA. Furthermore, these results highlight the broader potential of MLLMs, such as LLaVA, for classification tasks, particularly in MSA.

To facilitate further research and reproducibility, we have publicly released LLaVAC-7b\footnote{\url{https://huggingface.co/yacht/llavac-7b-msa}}, which was fine-tuned on the random dataset split.

\begin{table}[h!]
\centering
\scalebox{0.87}{
\begin{tabular}{lcc}
\hline
\textbf{Models} & \textbf{Acc} & $\textbf{F}_{1}$ \\ \hline
MultiSentiNet \citep{xu-etal-2017-multi}$\diamond$          & 63.27            & 59.12               \\ 
FENet-BERT \citep{jiang-etal-2020-fusion}$\diamond$       & 69.02               & 67.30                  \\ 
Se-MLNN \citep{cheema-etal-2021-fair}$\diamond$        & 75.33                & 73.76                   \\ 
VSA-PF \citep{chen-etal-2024-holistic}$\diamond$        & 75.58                & 74.48                   \\ 
LLaVAC (Ours)$\diamond$                                 & \textbf{76.48}       & \textbf{75.84}  \\
\hline
CMCN \citep{peng-etal-2022C-cross}$\star$          & 73.61            & 75.03                \\ 
CLMLF \citep{li-etal-2022-clmlf}$\star$       & 75.33               & 73.46                  \\ 
CBAM \citep{wang-etal-2023-exploring}$\star$        & 77.11                & 76.55                   \\ 
LLaVAC (Ours)$\star$                                 & \textbf{79.46}       & \textbf{79.00}                  \\ 
\hline
QMF \citep{zhang-etal-2023-provable}$\circ$        & 78.07                & -                   \\ 
LLaVAC (Ours)$\circ$      & \textbf{82.85}       & \textbf{82.03}                  \\
\hline
\end{tabular}}
\caption{Comparison of our model's results with previous works across different dataset splits. $\diamond$ indicates results using dataset splits similar to \citet{cheema-etal-2021-fair}, where both Acc and $\text{F}_{1}$ are averaged over the splits. $\star$ denotes results using the random dataset split, as outlined by \citet{wang-etal-2023-exploring}. $\circ$ represents results using the dataset split similar to \citet{zhang-etal-2023-provable}.}
\label{tab:main-result}
\end{table}

\subsection{Ablation Study}
We conducted an ablation study to assess the impact of using both unimodal and multimodal labels on model performance. Specifically, we modified the prompt to exclude unimodal labels, retaining only the multimodal label for fine-tuning LLaVA. This modification was intended to isolate the contribution of unimodal labels in the training process and examine their influence on MSA. Details of the modified prompt are provided in Appendix~\ref{appx:promt-unimodal}. The study was conducted using the random split of the MVSA-Single dataset.

Table~\ref{tab:ablation} presents a comparison between LLaVAC, fine-tuned with both unimodal and multimodal labels, and LLaVAC-MO, fine-tuned with only the multimodal label. LLaVAC achieves a 1.33\% higher accuracy and a 2.95\% higher weighted F1-score compared to LLaVAC-MO. The larger improvement in the weighted F1-score indicates enhanced balance between precision and recall across sentiment classes. These findings suggest that incorporating both unimodal and multimodal labels improves the model's overall performance in MSA.

\begin{table}[h!]
\centering
\scalebox{0.87}{
\begin{tabular}{lcc}
\hline
\textbf{Models} & \textbf{Acc} & $\textbf{F}_{1}$ \\ \hline
LLaVAC-MO$\star$   & 78.13       & 76.05                  \\ 
LLaVAC$\star$     & \textbf{79.46}       & \textbf{79.00}                 \\
\hline
\end{tabular}}
\caption{Ablation study comparing the accuracy and weighted F1-score of LLaVAC fine-tuned with only the multimodal label (LLaVAC-MO) and LLaVAC. $\star$ denotes results using the random dataset split.}
\label{tab:ablation}
\end{table}

\subsection{Discussion}
In this study, LLaVAC was evaluated on the MVSA-Single dataset. To ensure a robust assessment of its performance in sentiment analysis, we employed three distinct data processing procedures inspired by prior studies. These variations provided a comprehensive evaluation of the consistency and performance of the classifier across different splits.

Remarkably, our results show that LLaVAC outperformed all previous methods across all settings, achieving superior performance after fine-tuning for just one epoch and without relying on a validation set. This highlights the simplicity and effectiveness of our approach, demonstrating its ability to achieve state-of-the-art performance while maintaining a straightforward application process. The practical design of LLaVAC makes it a highly accessible and reliable solution for MSA.

\section{Conclusion}
We introduced LLaVAC, a method for constructing a classifier for multimodal sentiment analysis by fine-tuning LLaVA with a prompt design. LLaVAC processes both image and text modalities to generate sentiment labels, achieving state-of-the-art performance on the MVSA-Single dataset, surpassing existing methods in accuracy and weighted F1-score. Our findings demonstrate that LLaVA provides a strong foundation for building a multimodal sentiment classifier. Specifically, LLaVAC minimizes the need for complex manual feature engineering while maintaining robust performance.

\section{Limitations}

While LLaVAC achieves superior accuracy and weighted F1 scores compared to previous works, several limitations must be acknowledged.

\smallskip\noindent\textbf{Domain Applicability:}
This study evaluated LLaVAC on the MVSA-Single dataset, which originates from social networks. Its performance may depend on dataset-specific characteristics, and further research is needed to test its effectiveness across diverse datasets and domains.

\smallskip\noindent\textbf{Configuration Optimization:}
Factors such as model size, hyperparameters, and fine-tuning techniques may influence the effectiveness of LLaVAC. This study provides a baseline configuration, however, further experimentation is needed to identify optimal settings and improve performance.

\smallskip\noindent\textbf{Prompt Sensitivity:}
LLaVAC’s performance can vary with changes in prompt structure and phrasing, impacting output consistency. While this study employed a specific MSA-focused prompt, further research should explore how different prompt designs affect robustness and overall performance.

% Bibliography entries for the entire Anthology, followed by custom entries
%\bibliography{anthology,custom}
% Custom bibliography entries only
\bibliography{anthology_0,anthology_1,custom}

\appendix

\section{Prompt Template with Multimodal Label Only}
\label{appx:promt-unimodal}

\begin{figure}[htbp]
\centering
\begin{tcolorbox}[colframe=black, colback=white!0, boxrule=1pt, arc=0pt, boxsep=4pt, left=4pt, right=4pt, top=2pt, bottom=2pt]
\begin{algorithmic}
\STATE \textbf{Prompt:} 
\STATE Consider the following Image and Text:
\STATE Image: <image>
\STATE Text: RT @babeshawnmendes: "that was really energetic"
\STATE
\STATE Classify the Multimodal sentiment label (negative, neutral, positive).
\STATE Provide a short answer with 1 label for the Multimodal label.
\STATE
\STATE \textbf{Response:}
\STATE positive
\end{algorithmic}
\end{tcolorbox}
\caption{Example prompt with its response, including the multimodal label. <image> denotes the LLaVA-compatible image data, used for fine-tuning the model.}
\end{figure}

\end{document}